\documentclass[10pt,twocolumn,letterpaper]{article}

\usepackage{cvpr}              

\usepackage[dvipsnames]{xcolor}

\definecolor{cvprblue}{rgb}{0.21,0.49,0.74}
\usepackage[pagebackref,breaklinks,colorlinks,citecolor=cvprblue]{hyperref}
\usepackage{algorithmic}
\usepackage[ruled,linesnumbered]{algorithm2e}
\def\paperID{114514} 
\def\confName{CVPR}
\def\confYear{2024}

\begin{document}

\title{MetaScript: Few-Shot Handwritten Chinese Content Generation via Generative Adversarial Networks\\
\vspace{0.5em} \normalsize Project of AI3604 Computer Vision, 2023 Fall, SJTU}
\author{Jiazi Bu\footnotemark[1]\\
{\normalsize 521030910395}
\and
Qirui Li\footnotemark[1]\\
{\normalsize 521030910397}
\and
Kailing Wang\footnotemark[1]\\
{\normalsize 521030910356}
\and
Xiangyuan Xue\footnotemark[1]\\
{\normalsize 521030910387}
\and
Zhiyuan Zhang\footnotemark[1]\\
{\normalsize 521030910377}
}

\twocolumn[{%
\renewcommand\twocolumn[1][]{#1}%
\maketitle
\vspace{-1em}
\begin{center}
    \captionsetup{type=figure}
    \includegraphics[width=0.9\linewidth]{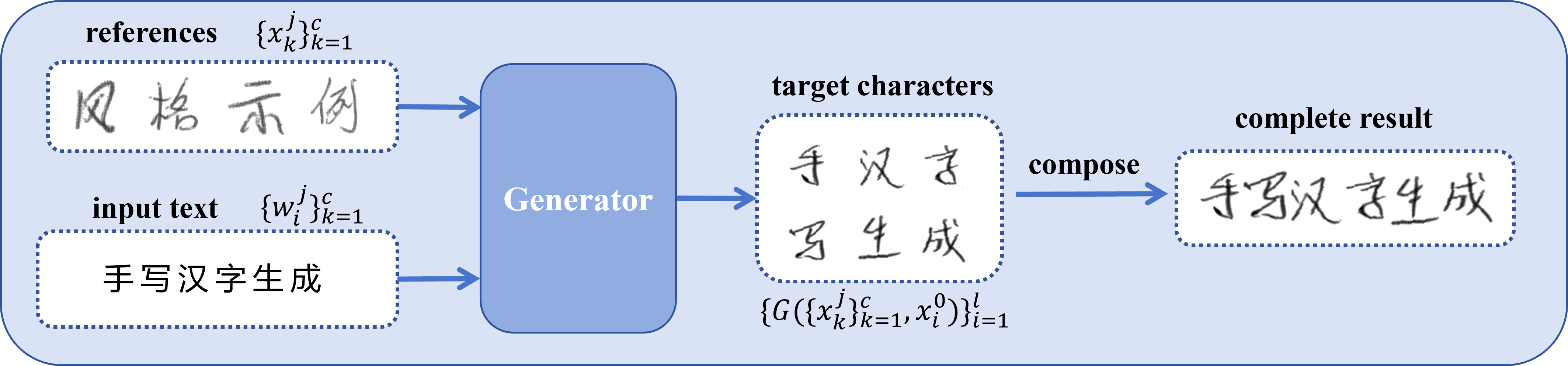}
    \captionof{figure}{The overall pipeline of this project. We design a system to generate handwritten Chinese contents within a few-shot setting. The system is composed of a generator and a composer. The generator is trained to generate handwritten Chinese characters given a structure template and some style references. The composer stitches the generated characters into a handwritten style content.}
    \label{fig:pipeline}
\end{center}
\vspace{1em}
}]

\renewcommand{\thefootnote}{*} \footnotetext[1]{The authors contributed equally to this project. The authors are arranged in alphabetical order by last name.}

\begin{abstract}
In this work, we propose MetaScript, a novel Chinese content generation system designed to address the diminishing presence of personal handwriting styles in the digital representation of Chinese characters. 
Our approach harnesses the power of few-shot learning to generate Chinese characters that not only retain the individual's unique handwriting style but also maintain the efficiency of digital typing. Trained on a diverse dataset of handwritten styles, MetaScript is adept at producing high-quality stylistic imitations from minimal style references and standard fonts. Our work demonstrates a practical solution to the challenges of digital typography in preserving the personal touch in written communication, particularly in the context of Chinese script. 
Notably, our system has demonstrated superior performance in various evaluations, including recognition accuracy, inception score, and Frechet inception distance. At the same time, the training conditions of our model are easy to meet and facilitate generalization to real applications.
Our code is available at \url{https://github.com/xxyQwQ/metascript}.
\end{abstract}  
\section{Introduction}
\label{sec:intro}

Chinese characters, utilized continuously for over six millennia by more than a quarter of the world's population, have been integral to education, employment, communication, and daily life in East Asia. The art of handwriting Chinese characters, transcends mere linguistic articulation, representing both the pinnacle of visual art and a medium for personal expression and cultivation \cite{personality}. There is an ancient Chinese saying, "seeing the character is like seeing the face." Throughout the process of personal growth, individuals develop distinct and characteristic handwriting styles. These styles can serve as symbols of one's identity. Since humanity's entry into the digital era, efficient yet characterless fixed fonts have supplanted handwritten text, eliminating the possibility of perceiving each other through the written word. This shift has engendered a sense of detachment from handwriting, significantly diminishing the personalization and warmth in textual communication. 

To address this issue, we aim to devise a method that retains the individual's handwriting style while also harnessing the efficiency afforded by typing. However, Chinese characters, numbering over 100,000 distinct ideograms with diverse glyph structures, lack standardized stroke units. Generating handwritten Chinese characters in a specific style is challenging through the naive structure-based approaches. Although there have been attempts to utilize the stroke decomposition of Chinese characters \cite{giffchi1}, combined with vision transformer techniques for morphological imitation, such methods require a substantial reference character set or a highly complex character decomposition tree to generate new styles. They require extensive linguistic processing, significant storage consumption, and complex search processes, rendering them impractical for everyday use.

Consequently, we introduce MetaScript, a novel approach designed for generating a large number of Chinese characters in a consistent style using few-shot learning. Our method is trained on a handwritten dataset encompassing a variety of styles and a substantial quantity of Chinese characters. It is capable of producing high-quality stylistic imitations of a vast array of text, utilizing only a few style references and standard fonts as structural information. This approach effectively bridges the gap between the personalized nuances of handwriting and the efficiency of digital text generation.

Our work has a multitude of applications. For instance, it can serve as a straightforward alternative to traditional font generation methods, facilitating the effortless creation of personalized fonts. Our approach is capable of producing unique glyph designs, which can be instrumental in artistic and visual media design. With its low computational demands and real-time inference capabilities, our work can be integrated with large language models to generate responses in personalized fonts, thereby enriching the user experience.

We summarize our contributions in three key aspects:

\begin{itemize}
\item \textbf{Innovative Few-Shot Learning Model:} MetaScript employs a few-shot learning framework that enables the model to learn and replicate a specific handwriting style from a minimal set of examples. This significantly reduces the need for extensive training data, making the system more efficient and adaptable to individual styles.

\item \textbf{Integration of Structural and Stylistic Elements:} Our approach uniquely combines the structural integrity of standard Chinese fonts with the stylistic elements of individual handwriting using structure and style encoders. This integration ensures that the generated characters are not only stylistically consistent but also maintain the legibility and structural accuracy essential for Chinese script.

\item \textbf{Scalability and Efficiency:} The MetaScript system is designed to be scalable, capable of handling the generation of a vast number of characters without a proportional increase in computational resources or storage. This scalability is crucial given the extensive number of ideograms in the Chinese language and is a significant advancement over previous methods that required substantial storage and processing power.

\end{itemize}

\section{Related Work}
\label{sec:related_work}



\subsection{Generative Adversarial Networks}
Generative Adversarial Networks (GANs) have been a hot research topic in the past 10 years \cite{ganreview}. For example, there are more than 22,900 papers related to GAN in 2023, that is: more than 2.5 papers per hour. 
GANs involve two parts: a generator that creates data and a discriminator that distinguishes between generated and real data. They are trained through a minimax optimization to reach a Nash equilibrium \cite{nash}, where the generator effectively replicates the real data distribution.
Thanks to the numerous works to enhance the objective function \cite{cyclegan, rgans}, structure \cite{stylegan, gman, madgan, cogan}, and transfer learning ability \cite{cyclegan, stylegan, fcan}, of GANs, GANs now have a multitude of applications across various domains, such as Super-resolution \cite{ganphoto, esrgan, cicgan, tgan}, Image synthesis and manipulation \cite{sgan, vgan}, detection and video processing and NLP tasks \cite{mogren, seqgan}.

\subsection{English Handwriting Generation}
The generation of handwritten text represents a historically longstanding and classic task \cite{hanreview}. Early in 2007, Gangadhar et al. attempted to use Oscillatory neural networks to generate handwriting \cite{2007}. Some works used deep recurrent neural networks \cite{rehan1, rehan2, rehan3} to ensure enhanced consistency in the generated outcomes is imperative. Kanda et al.  \cite{rlhan} proposes the use of reinforcement learning to evolve a rigorous future evaluation in traning. Other works imployed GANs to perform this task \cite{ganhan1, ganhan2, ganhan3}.

\subsection{CJK Character Generation}
Unlike some alphabetic languages that can generate extensive text using a limited number of templates, CJK (Chinese, Japanese, and Korean) languages are characterized by their abundant and structurally complex characters, precluding the possibility of generation through a minimal set of templates. The task of CJK (especially Chinese) character generation can be traced back to a period when computers had not yet become widespread in China \cite{chigenori}. Early methods decomposed characters into components or strokes \cite{chier1, chier2}. Later studies applied deep learning. Some of them used GANs or similar structure to transfer certain style onto stereotypes \cite{chigan1, chigan2, chigan3, wnet}. Later advancement used Diffusion models to generate font sets \cite{giffchi1, giffchi2, giffchi3}.
\section{Method}
\label{sec:method}

\subsection{Overall Pipeline}

Suppose the dataset $\mathcal{D}$ contains $n$ types of Chinese characters and $m$ writers totally. Let $x_i^j$ denote the character with the $i$-th type written by the $j$-th writer, i.e. the script, where $i \in \{1, 2, \dots, n\}$ and $j \in \{1, 2, \dots, m\}$. Specifically, let $x_i^0$ denote the character with the $i$-th type rendered from a standard font, i.e. the prototype. The proposed character generator $G$ is trained to follow the style of the references while keeping the structure of the templates. To be exact, given some references $\{x_k^j\}_{k=1}^c$ and a template $x_i^0$, the generated result $G(\{x_k^j\}_{k=1}^c, x_i^0)$ should be similar to $x_i^j$, which inherits the structure of the $i$-th type and the style of the $j$-th writer. We expect the character generator $G$ can generalize well to the unseen references.

We utilize the adversarial learning paradigm to train the character generator $G$. Specifically, we introduce a multi-scale discriminator $D$ to distinguish the generated character $G(\{x_k^j\}_{k=1}^c, x_i^0)$ from the real character $x_i^j$. The discriminator $D$ is also trained to predict the type and writer of the given character $x$. We expect that the discriminator $D$ can encourage the generator $G$ to learn how to generate plausible characters.

Based on the character generator $G$, we can build a Chinese content generating system $S$. Given some style references $\{x_k\}_{k=1}^c$ and a content text $\{w_i\}_{i=1}^l$, the system $S$ retrieves the corresponding templates $\{x_{w_i}^0\}_{i=1}^l$, generates the target characters $\{G(\{x_k\}_{k=1}^c, x_{w_i}^0)\}_{i=1}^l$, and composes them into the complete result. Such pipeline is defined as few-shot handwritten Chinese character generation.

\begin{figure*}[t]
    \centering
    \includegraphics[width=1.0\linewidth]{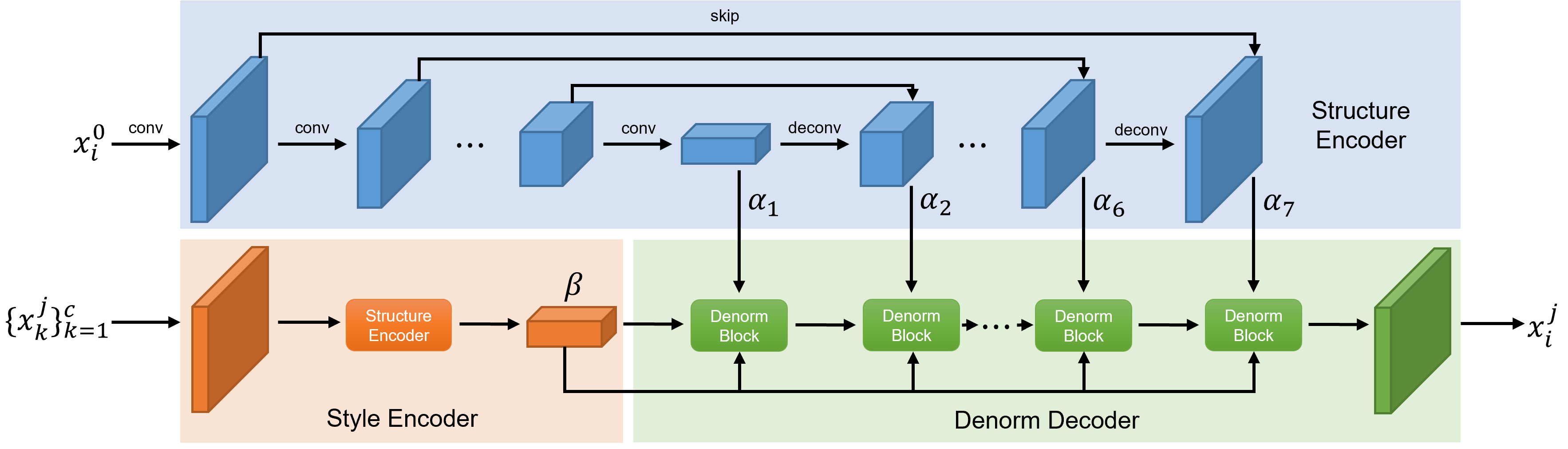}
    \caption{The overview of the proposed character generator $G$. The generator $G$ mainly contains three modules: a structure encoder $E_{\alpha}$, a style encoder $E_{\beta}$, and a denormalization decoder $D_{\gamma}$. The structure encoder $E_{\alpha}$ applies the U-Net architecture. The style encoder $E_{\beta}$ applies the ResNet-18 architecture. The denormalization decoder $D_{\gamma}$ is composed of $7$ cascaded denormalization blocks.}
    \label{fig:generator_overview}
\end{figure*}

\begin{figure*}[t]
    \centering
    \includegraphics[width=0.9\linewidth]{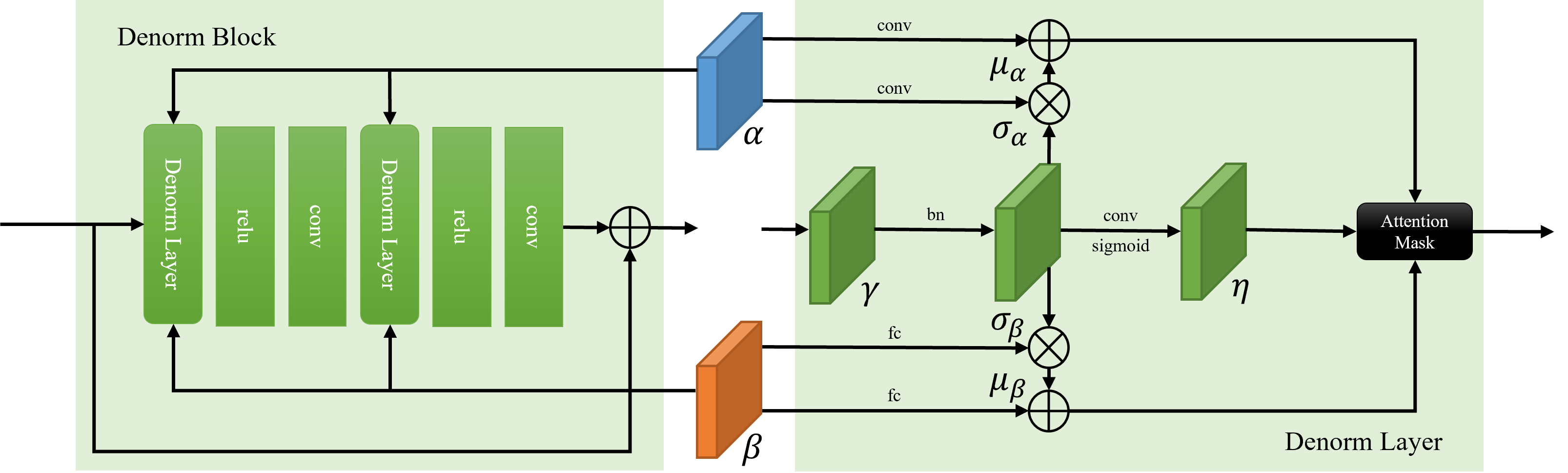}
    \caption{The detailed structure of the denormalization block and the denormalization layer. The denormalization layer is composed of a normalization step, a denormalization step and an attention mechanism. The denormalization block is composed of a skip connection and two identical layers, each of which contains a denormalization layer, an activation layer and a convolution layer.}
    \label{fig:generator_detail}
\end{figure*}

\subsection{Character Generator}

Inspired by previous generative works \cite{faceshifter,stylegan,stylegan2}, our proposed character generator $G$ mainly contains three modules: a structure encoder $E_{\alpha}$, a style encoder $E_{\beta}$, and a denormalization decoder $D_{\gamma}$. The structure encoder $E_{\alpha}$ extracts the structure information $\alpha_1,\alpha_2,\dots,\alpha_7$ from the template $x_i^0$. The style encoder $E_{\beta}$ extracts the style information $\beta$ from the references $\{x_k^j\}_{k=1}^c$. Then the denormalization decoder $D_{\gamma}$ combines the structure and style information to generate the target character $x_i^j$. The overview of the proposed character generator is shown in Figure \ref{fig:generator_overview}. The loss function will be described in Equation \ref{eq:loss_overall_generator} and \ref{eq:loss_overall_discriminator}.

\noindent \textbf{Structure Encoder.} The structure encoder $E_{\alpha}$ applies the U-Net \cite{unet} architecture, which includes $6$ down-sampling blocks and $6$ up-sampling blocks, extracting $7$ feature maps $\alpha_1,\alpha_2,\dots,\alpha_7$ with different scales from the template $x_i^0$, as shown in the blue part of Figure \ref{fig:generator_overview}.
\begin{equation}
    E_{\alpha}(x_i^0) = \{\alpha_1,\alpha_2,\dots,\alpha_7\}.
    \label{eq:structure_encoder}
\end{equation}
Each block is composes of a $4 \times 4$ convolution layer with stride $2$, a normalization layer, and an activation layer. There are skip connections between feature maps with the same scale. The structure encoder $E_{\alpha}$ is trained in a self-supervised manner, which expects the structure of the generated character $G(\{x_k^j\}_{k=1}^c, x_i^0)$ to be the same with that of the template $x_i^0$. The loss function will be described in Equation \ref{eq:loss_structure}.

\noindent \textbf{Style Encoder.} The style information should be as concise as a dense feature vector. Therefore, we apply the ResNet-18 \cite{resnet} architecture in the style encoder $E_{\beta}$ with input channels modified to $c$ and a linear layer added at the end. The style encoder $E_{\beta}$ extracts a $512$-dimensional feature vector $\beta$ from the references $\{x_k^j\}_{k=1}^c$ as the style information, as shown in the orange part of Figure \ref{fig:generator_overview}.
\begin{equation}
    E_{\beta}(\{x_k^j\}_{k=1}^c) = \beta.
    \label{eq:style_encoder}
\end{equation}
Similar to the structure encoder $E_{\alpha}$, the style encoder $E_{\beta}$ is trained in a self-supervised manner, which expects the style of the generated character $G(\{x_k^j\}_{k=1}^c, x_i^0)$ to be the same with that of the references $\{x_k^j\}_{k=1}^c$. The loss function will be described in Equation \ref{eq:loss_style}.

\noindent \textbf{Denormalization Decoder.} Both the structure information $\alpha_1,\alpha_2,\dots,\alpha_7$ and the style information $\beta$ will be fed into the denormalization decoder $D_{\gamma}$, which is composed of $7$ cascaded denormalization blocks $D_{\gamma}^1,D_{\gamma}^2,\dots,D_{\gamma}^7$, as shown in the green part of Figure \ref{fig:generator_overview}.
\begin{equation}
    D_{\gamma}(\{\alpha_i\}_{i=1}^7, \beta) = G(\{x_k^j\}_{k=1}^c, x_i^0).
    \label{eq:denormalization_decoder}
\end{equation}
The output of the denormalization decoder $D_{\gamma}$ is the generated character $G(\{x_k^j\}_{k=1}^c, x_i^0)$, which will be directly supervised by the ground truth $x_i^j$.

The detailed structure of the denormalization block is shown in Figure \ref{fig:generator_detail}. The denormalization block is composed of two identical layers, each of which contains a denormalization layer, an activation layer and a $3 \times 3$ convolution layer. There is also a skip connection between the input and output of the denormalization block, which follows the classical residual learning strategy \cite{resnet}.
\begin{equation}
    D_{\gamma}^i(\alpha_i, \beta, \gamma_{i-1}) = \gamma_i,
    \label{eq:denormalization_block}
\end{equation}
where $i\in \{1, 2, \dots, 7\}$. Specifically, we state that $\gamma_0=\beta$ and $\gamma_7=G(\{x_k^j\}_{k=1}^c, x_i^0)$ for simplicity.

The detailed structure of the denormalization layer is also shown in Figure \ref{fig:generator_detail}. The denormalization layer follows the design of Adaptive Instance Normalization (AdaIN) \cite{stylegan}, which is composed of a normalization step and a denormalization step. Inspired by  \cite{faceshifter}, we also introduce an attention mechanism to fuse different feature maps softly. To be exact, the input feature map $\gamma$ is first normalized in the channel dimension.
\begin{equation}
    \bar{\gamma} = \frac{\gamma - \mu_{\gamma}}{\sigma_{\gamma}},
    \label{eq:denormalization_layer_norm}
\end{equation}
where $\mu_{\gamma}$ and $\sigma_{\gamma}$ are the mean and standard deviation of the input feature map $\gamma$ respectively.

Then the structure feature map $\alpha$ will be fed into a $1 \times 1$ convolution layer to predict the mean $\mu_{\alpha}$ and standard deviation $\sigma_{\alpha}$ of the normalized feature map $\bar{\gamma}$ for warping.
\begin{equation}
    \hat{\alpha} = \sigma_{\alpha} \times \bar{\gamma} + \mu_{\alpha},
    \label{eq:denormalization_layer_warp_alpha}
\end{equation}
where $\hat{\alpha}$ is the predicted structure feature map.

The style feature vector $\beta$ will be fed into a linear layer to predict the mean $\mu_{\beta}$ and standard deviation $\sigma_{\beta}$ of the normalized feature map $\bar{\gamma}$ for warping.
\begin{equation}
    \hat{\beta} = \sigma_{\beta} \times \bar{\gamma} + \mu_{\beta},
    \label{eq:denormalization_layer_warp_beta}
\end{equation}
where $\hat{\beta}$ is the predicted style feature map.

In addition, the normalized feature map $\bar{\gamma}$ will be fed into a $1 \times 1$ convolution layer and a sigmoid activation layer to form the attention map $\eta$, which is used as a weighted mask to fuse the feature maps $\hat{\alpha}$ and $\hat{\beta}$.
\begin{equation}
    \hat{\gamma} = (1 - \eta) \times \hat{\alpha} + \eta \times \hat{\beta},
    \label{eq:denormalization_layer_attention}
\end{equation}
where $\hat{\gamma}$ is the output feature map. Finally, the denormalization layer completes the entire process.

The key idea of the denormalization layer is to adaptively adjust the effective regions of the structure feature map and the style feature map, so that the generated character can inherit the structure of the template and the style of the references. Compared with the AdaIN \cite{stylegan}, the denormalization layer can fuse the feature maps of arbitrary styles instead of pairwise exchanging, which shows better flexibility and diversity in character generation.

\subsection{Multi-scale Discriminator}

\begin{figure}[htbp]
    \centering
    \includegraphics[width=0.9\linewidth]{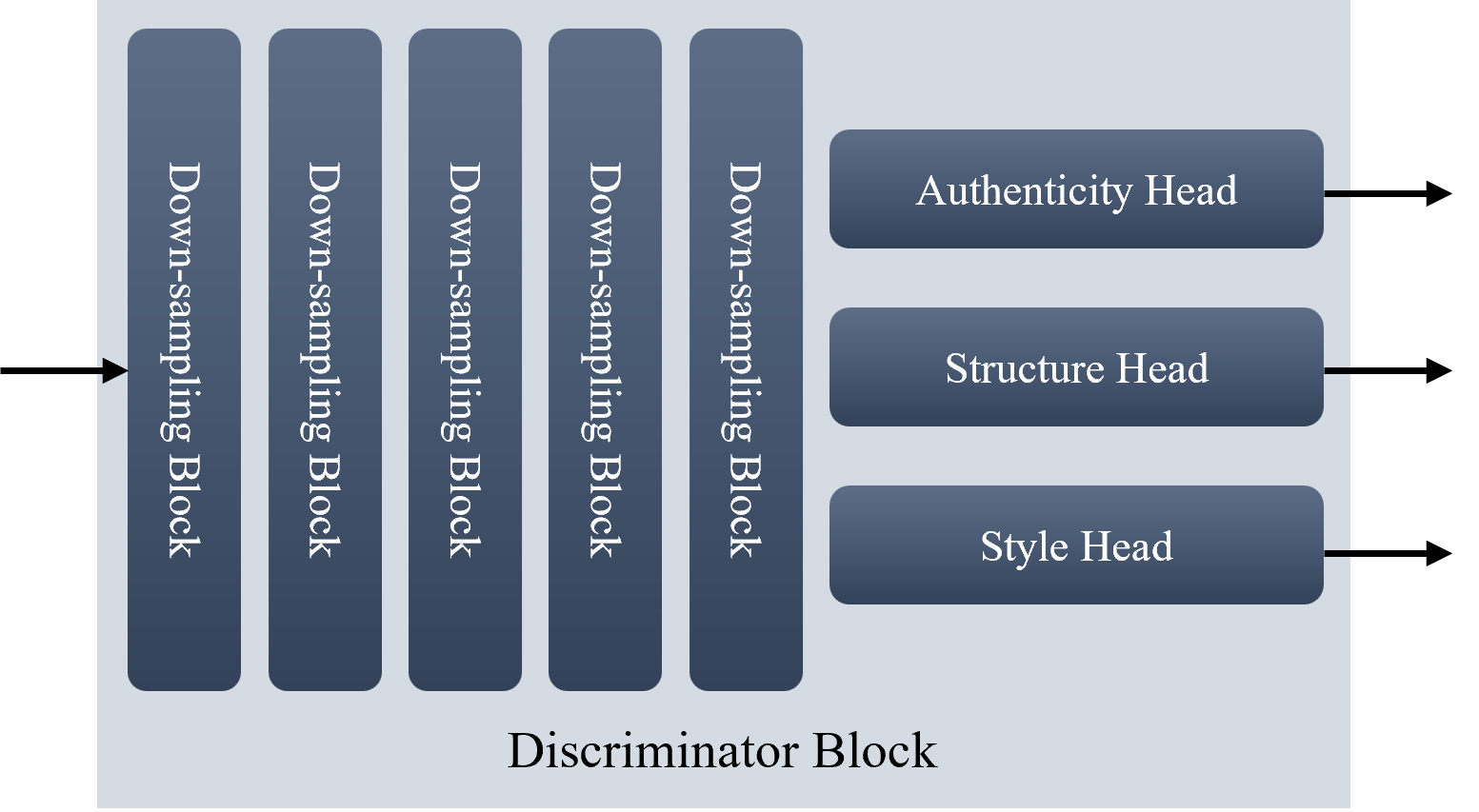}
    \caption{The structure of the discriminator block. The discriminator block is composed of $5$ down-sampling blocks and $3$ classification heads as a typical convolutional neuron network.}
    \label{fig:discriminator_block}
\end{figure}

The discriminator block $D^i$ follows the traditional convolutional neuron network paradigm, which is composed of $5$ down-sampling blocks and $3$ classification heads, as shown in Figure \ref{fig:discriminator_block}. Each down-sampling block is composed of a $4 \times 4$ convolution layer with stride $2$, a normalization layer, and an activation layer, which is exactly the same as that of the structure encoder $E_{\alpha}$. Each classification head contains a single linear layer to predict the probability distribution from the extracted feature map.
\begin{equation}
    D^i(x) = (D_{\phi}^i(x), D_{\alpha}^i(x), D_{\beta}^i(x)) = (\hat{y}_{\phi}^i, \hat{y}_{\alpha}^i, \hat{y}_{\beta}^i),
    \label{eq:discriminator_block}
\end{equation}
where $D^i$ represents the $i$-th discriminator block, $\hat{y}_{\phi}^i$ represents the predicted authenticity, $\hat{y}_{\alpha}^i$ represents the predicted type, and $\hat{y}_{\beta}^i$ represents the predicted writer. We expect that the discriminator block can extract useful features to distinguish the authenticity, type and writer all together. The fundamental idea is to force the generator to learn the correct structure and style features, instead of simply cheating the discriminator. The loss function will be described in Equation \ref{eq:loss_adversarial_generator}, \ref{eq:loss_adversarial_discriminator}, \ref{eq:loss_classification_generator} and \ref{eq:loss_classification_discriminator}.

\begin{figure}[htbp]
    \centering
    \includegraphics[width=1.0\linewidth]{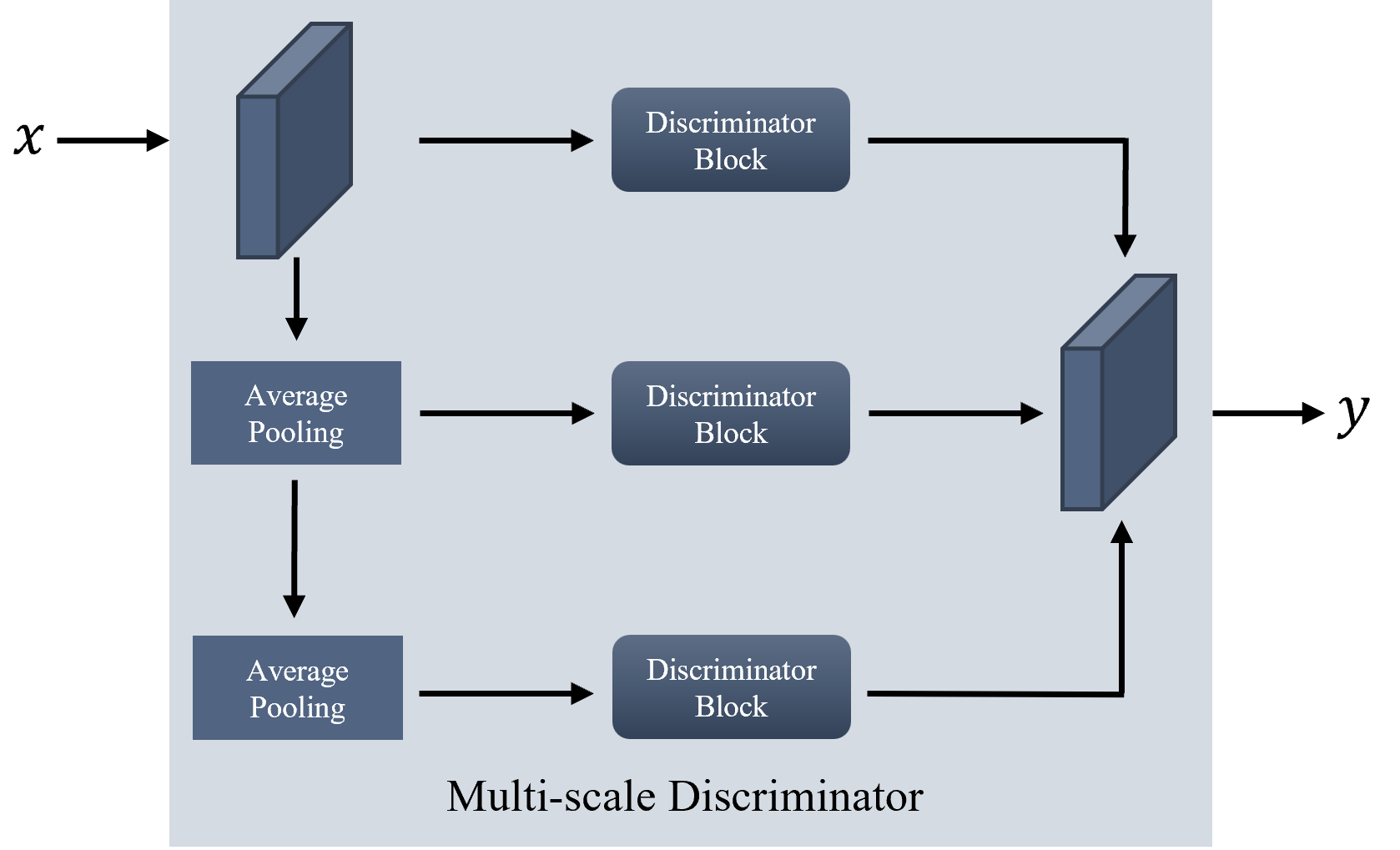}
    \caption{The overview of the multi-scale discriminator $D$. The discriminator $D$ is composed of $2$ average pooling layers and $3$ discriminator blocks $D^1$, $D^2$ and $D^3$. The input character $x$ will be down-sampled by the average pooling layers to form $3$ different scales and fed into the corresponding discriminator blocks.}
    \label{fig:discriminator_multiscale}
\end{figure}

Inspired by \cite{pix2pixhd}, we apply a multi-scale discriminator $D$ to enhance the performance of the discriminator, as shown in Figure \ref{fig:discriminator_multiscale}. The multi-scale discriminator $D$ is composed of $2$ average pooling layers and $3$ discriminator blocks $D^1$, $D^2$ and $D^3$. The input character $x$ will be down-sampled by the average pooling layers to form $3$ different scales, so the corresponding discriminator blocks can evaluate the input character $x$ from different perspectives and perform better supervision. Previous works \cite{pix2pixhd,faceshifter} have shown that the multi-scale discriminator can effectively improve the quality of the generated results, especially in the high-resolution tasks.
\begin{equation}
    D(x) = \{D^1(x), D^2(x'), D^3(x'')\},
    \label{eq:discriminator_multiscale}
\end{equation}
where $x'$ and $x''$ represent the input character $x$ down-sampled once and twice respectively. The loss function will be described in Equation \ref{eq:loss_overall_generator} and \ref{eq:loss_overall_discriminator}.

\subsection{Training Objective}

We utilize adversarial learning to train the character generator $G$ and the multi-scale discriminator $D$ and introduce $5$ kinds of loss functions: adversarial loss $\mathcal{L}_{adv}$, classification loss $\mathcal{L}_{cls}$, structure loss $\mathcal{L}_{str}$, style loss $\mathcal{L}_{sty}$ and reconstruction loss $\mathcal{L}_{rec}$. The overall loss function is a weighted sum over them. Formally, for the generator, the overall loss function is defined as
\begin{equation}
    \begin{aligned}
        \mathcal{L}_{all}^{G} &= \lambda_{adv}^G \mathcal{L}_{adv}^G + \lambda_{cls}^G \mathcal{L}_{cls}^G \\
        &+ \lambda_{str}^G \mathcal{L}_{str}^G + \lambda_{sty}^G \mathcal{L}_{sty}^G + \lambda_{rec}^G \mathcal{L}_{rec}^G,
    \end{aligned}
    \label{eq:loss_overall_generator}
\end{equation}
and for the discriminator it is defined as
\begin{equation}
    \begin{aligned}
        \mathcal{L}_{all}^{D} &= \lambda_{adv}^D \mathcal{L}_{adv}^D + \lambda_{cls}^D \mathcal{L}_{cls}^D,
    \end{aligned}
    \label{eq:loss_overall_discriminator}
\end{equation}
where $\lambda_{adv}^G$, $\lambda_{cls}^G$, $\lambda_{str}^G$, $\lambda_{sty}^G$, $\lambda_{rec}^G$, $\lambda_{adv}^D$ and $\lambda_{cls}^D$ are the hyperparameters to balance the loss functions.

\noindent \textbf{Adversarial Loss.} The adversarial loss is to train the discriminator $D$ to distinguish the generated character $G(\{x_k^j\}_{k=1}^c, x_i^0)$ from the real character $x_i^j$, which indirectly encourages the generator $G$ to generate more plausible characters. Binary cross entropy is applied as the adversarial loss. Formally, for the generator, the adversarial loss is defined as
\begin{equation}
    \begin{aligned}
        \mathcal{L}_{adv}^G = &- \sum_{s=1}^3 \log D_{\phi}^s (G(\{x_k^j\}_{k=1}^c, x_i^0)), \\
    \end{aligned}
    \label{eq:loss_adversarial_generator}
\end{equation}
and for the discriminator it is defined as
\begin{equation}
    \begin{aligned}
        \mathcal{L}_{adv}^D = &- \sum_{s=1}^3 \log D_{\phi}^s (x_i^j) \\
        &- \sum_{s=1}^3 \log [1 - D_{\phi}^s (G(\{x_k^j\}_{k=1}^c, x_i^0))].
    \end{aligned}
    \label{eq:loss_adversarial_discriminator}
\end{equation}

\noindent \textbf{Classification Loss.} The classification loss is to train the discriminator $D$ to precisely predict the type and writer of the given character $x$. Different from the adversarial loss, the generator $G$ is also trained to minimize the classification loss, which indirectly encourages the generator $G$ to generate characters with accurate structure and style. Cross entropy is applied as the classification loss. Formally, for the generator, the classification loss is defined as
\begin{equation}
    \begin{aligned}
        \mathcal{L}_{cls}^G = &- \sum_{s=1}^3 \log D_{\alpha}^s (G(\{x_k^j\}_{k=1}^c, x_i^0)) \\
        &- \sum_{s=1}^3 \log D_{\beta}^s (G(\{x_k^j\}_{k=1}^c, x_i^0)),
    \end{aligned}
    \label{eq:loss_classification_generator}
\end{equation}
and for the discriminator it is defined as
\begin{equation}
    \begin{aligned}
        \mathcal{L}_{cls}^D = &- \sum_{s=1}^3 \log D_{\alpha}^s (x_i^j) - \sum_{s=1}^3 \log D_{\beta}^s (x_i^j) \\
        &- \sum_{s=1}^3 \log [D_{\alpha}^s (G(\{x_k^j\}_{k=1}^c, x_i^0))] \\
        &- \sum_{s=1}^3 \log [D_{\beta}^s (G(\{x_k^j\}_{k=1}^c, x_i^0))].
    \end{aligned}
    \label{eq:loss_classification_discriminator}
\end{equation}
We should note that the classification loss is indispensable, which introduces effective supervision to prevent the generator from simply generating meaningless characters to cheat the discriminator. We will show how the classification loss solves the problem of mode collapse in Section \ref{sec:experiments}.

\noindent \textbf{Structure Loss.} Intuitively, we expect that the generated character $G(\{x_k^j\}_{k=1}^c, x_i^0)$ can inherit the structure of the template $x_i^0$. Therefore, The feature maps $\alpha_1,\alpha_2,\dots,\alpha_7$ extracted by the structure encoder $E_{\alpha}$ should be invariant. The structure loss not only encourages the generator $G$ to generate the correct structure, but also encourages the structure encoder $E_{\alpha}$ to extract valid structure features. The structure loss for the generator is formally defined as
\begin{equation}
    \mathcal{L}_{str}^G = \frac{1}{2} \| E_{\alpha}(G(\{x_k^j\}_{k=1}^c, x_i^0)) - E_{\alpha}(x_i^0) \|_2^2.
    \label{eq:loss_structure}
\end{equation}

\noindent \textbf{Style Loss.} We also expect that the generated character $G(\{x_k^j\}_{k=1}^c, x_i^0)$ can follow the style of the references $\{x_k^j\}_{k=1}^c$. Therefore, the feature vector $\beta$ extracted by the style encoder $E_{\beta}$ should be invariant. The style loss not only encourages the generator $G$ to generate the correct style, but also encourages the style encoder $E_{\beta}$ to extract valid style features. The style loss for the generator is formally defined as
\begin{equation}
    \mathcal{L}_{sty}^G = \frac{1}{2} \| E_{\beta}(G(\{x_k^j\}_{k=1}^c, x_i^0)) - E_{\beta}(\{x_k^j\}_{k=1}^c) \|_2^2.
    \label{eq:loss_style}
\end{equation}

\noindent \textbf{Reconstruction Loss.} The reconstruction loss represents the pixel-wise difference between the generated character $G(\{x_k^j\}_{k=1}^c, x_i^0)$ and the ground truth $x_i^j$, which is a direct supervision to the generator $G$. $L_1$ norm is applied as the reconstruction loss. The reconstruction loss for the generator is formally defined as
\begin{equation}
    \mathcal{L}_{rec}^G = \| G(\{x_k^j\}_{k=1}^c, x_i^0) - x_i^j \|_1.
    \label{eq:loss_reconstruction}
\end{equation}
We should note that the reconstruction loss is not to force the generated character to be exactly the same with the ground truth because such constraint will limit the diversity of the generated characters. Instead, the reconstruction loss should be controlled to a reasonable range.

\subsection{Content Composition}
We develop a typesetting tool, Typewriter, for reorganizing output characters into a complete result. For a given text and some style references, we generate an individual character for each character in the input text using the methods described in the preceding subsections. To better and more intuitively display the effectiveness of our method in generating Chinese character content, we first apply a random transformation to the generated character images, mimicking the alignment effect of human handwriting. Subsequently, they are arranged into a complete image. The details are shown in Algorithm \ref{algo:tprt}.

\begin{algorithm}[ht]
\small
\caption{Typewriter Procedure}\label{algo:tprt}
\SetKwInOut{Input}{Input}
\SetKwInOut{Output}{Output}
\SetKwComment{Comment}{/* }{ */}
\SetKwInOut{Define}{Functions}
\Input{style references $\{x_k\}_{k=1}^c$, content text $\{w_i\}_{i=1}^l$, expected character size $size_c$, expected line width $width_l$}
\Output{arranged handwritten content image $I_a$}
\BlankLine
Initialize: $cursor \leftarrow$ position $0$, $I_a\leftarrow$ empty image\;
\For{$w_i$ in $\{w_i\}_{i=1}^l$}{
    \uIf{$w_i$ is line break}{
        move $cursor$ to next line\;
        continue\;
    }
    \uElseIf{$w_i$ is space}{
        $g_i\leftarrow$ empty space with $\frac{size_c}{2}$ width\;
    }
    \uElseIf{$w_i$ is character}{
        generate $g_i\leftarrow G(\{x_k\}_{k=1}^c, x_{w_i}^0)$ with $size_c$\;
    }
    \ElseIf{$w_i$ is punctuation}{
        $g_i\leftarrow$ punctuation template\;
    }
    apply random transformation on $g_i$\;
    plot $g_i$ at $cursor$ in $I_a$ and update $cursor$\;
    \If{$cursor > width_l$}{
        move $cursor$ to next line\;
    }
}
return handwritten content image $I_a$
\end{algorithm}
\section{Experiments}
\label{sec:experiments}

\begin{figure*}[t]
    \centering
    \includegraphics[width=0.9\linewidth]{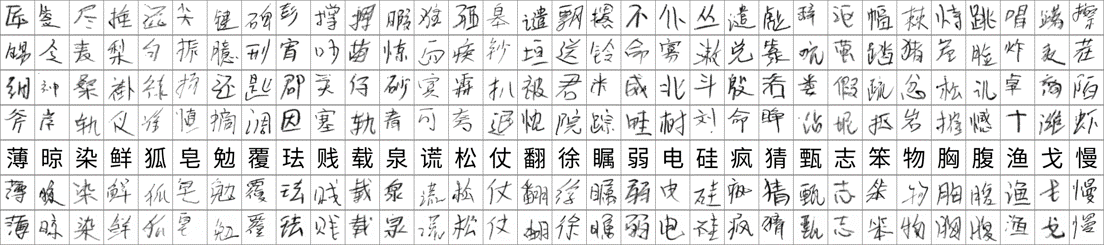}
    \caption{Characters synthesized by the generator with $4$ references trained for $100\text{k}$ iterations. The first $4$ rows are the style references. The $5$th row is the structure template. The $6$th row is the ground truth. The last row is the generated character.}
    \label{fig:demo_character}
\end{figure*}

\begin{figure*}[t]
    \centering
    \includegraphics[width=0.9\linewidth]{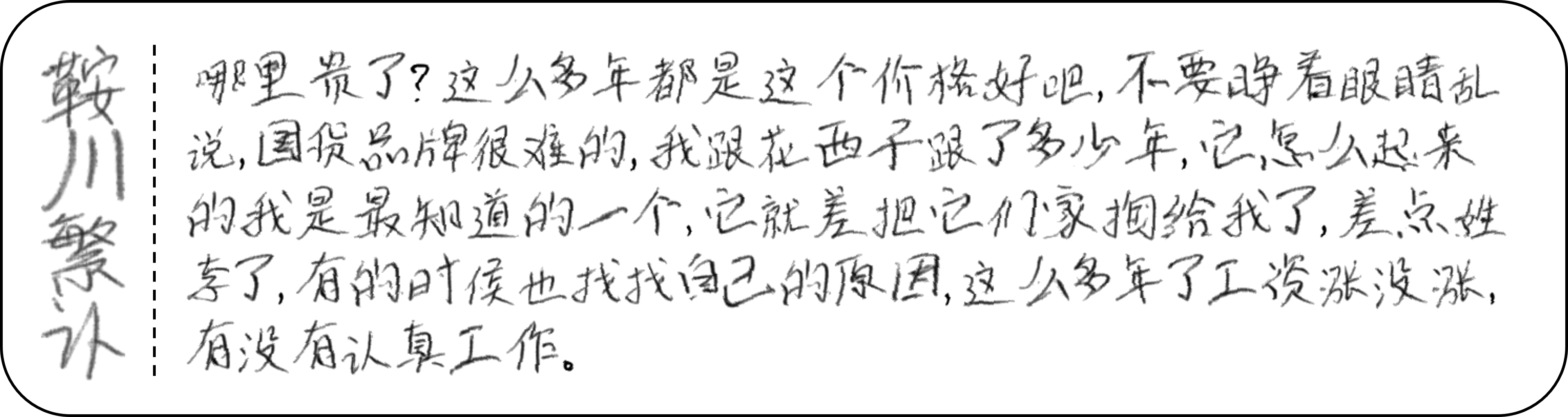}
    \caption{Content generated by the system, which is built based on the generator and cascaded with the type writter. The left hand side shows the style references. The right hand side shows the generated results.}
    \label{fig:demo_content}
\end{figure*}

\subsection{Experiment Setup}

We finetune the hyperparameters, train the model and build the system. Some results are shown in Figure \ref{fig:demo_character} and \ref{fig:demo_content}. The training details are as follows.

\noindent \textbf{Dataset.} The training set is build based on the CASIA-HWDB-1.1 dataset \cite{casia}. We select approximately $1\text{M}$ script images from $300$ writers, including $3755$ common Chinese characters. By preprocessing, the script images are resized into $128\times 128$ resolution, named by the types and classified by the writers, which will be used as style references. We also render the prototype images of the characters with the same resolution from the Source Han Sans CN font, which will be used as structure templates.

\noindent \textbf{Implementation details.} The model is trained for $100\text{k}$ iterations with a batch size of $32$. We use the Adam optimizer for training. The learning rate is set to $0.0001$ for both the generator and the discriminator. The weights of the loss function are $\lambda_{\text{adv}}^G=1$, $\lambda_{\text{cls}}^G=1$, $\lambda_{\text{str}}^G=0.5$, $\lambda_{\text{sty}}^G=0.1$, $\lambda_{\text{rec}}^G=20$, $\lambda_{\text{adv}}^D=1$ and $\lambda_{\text{cls}}^D=1$.

\noindent \textbf{Evaluation metrics.} To evaluate the quality and diversity of the generated results, we apply recognition accuracy (RA), inception score (IS) \cite{is} and Frechet inception distance (FID) \cite{fid} as the evaluation metrics. We use an EfficientNetV2 model \cite{efficientnetv2} pretrained on the CASIA-HWDB-1.1 dataset as the classifier to extract high-level features and perform low-level predictions for computation. The test set is manually selected from the CASIA-HWDB-1.0 dataset, which contains $1000$ characters written by $10$ writers.

\subsection{Ablation Studies}

We conduct two separate ablation studies to investigate the best number of references and verify the effectiveness of each loss term.

\noindent \textbf{Reference number.} To figure out how the number of references affects the performance of the model, we train the model with $1$, $2$, $4$ and $8$ references for $100\text{k}$ iterations respectively. The results are shown in Table \ref{tab:ref_num}.

\begin{table}[htbp]
\centering
\caption{Different reference numbers exhibit variations in model performance, including recognition accuracy (RA), inception score (IS) and Frechet inception distance (FID).}
\label{tab:ref_num}
\begin{tabular}{cccc}
\toprule
Reference & RA$\uparrow$ & IS$\uparrow$ & FID$\downarrow$ \\
\midrule
$1$ & $79.8\%$ & $55.543$ & $197.950$ \\
$2$ & $77.5\%$ & $58.524$ & $190.193$ \\
$4$ & $\mathbf{81.9}\%$ & $58.172$ & $\mathbf{187.365}$ \\
$8$ & $76.2\%$ & $\mathbf{58.598}$ & $188.411$ \\
\bottomrule
\end{tabular}
\end{table}

We can see that in general, as the number of references increases, the inception score increases and the Frechet inception distance decreases, which indicates better authenticity and diversity of the generated results. This is reasonable because more references provide more sufficient style information for the generator. However, the recognition accuracy shows strong fluctuations and achieves the best performance with $4$ references. Therefore, we prefer to use $4$ references in practice.

\noindent \textbf{Loss function.} Another question is that whether each loss term is necessary for training the model. To answer this question, we train the model with $4$ references for $100\text{k}$ iterations, but each loss term is removed respectively. The results are shown in Table \ref{tab:loss_weight}.

\begin{table}[htbp]
\centering
\caption{Changes in model performance with different loss terms removed, including adversarial loss ($\mathcal{L}_{adv}$), classification loss ($\mathcal{L}_{cls}$), structure loss ($\mathcal{L}_{str}$), style loss ($\mathcal{L}_{sty}$) and reconstruction loss ($\mathcal{L}_{rec}$), compared to the baseline model ($\mathcal{L}_{all}$).}
\label{tab:loss_weight}
\begin{tabular}{cccc}
\toprule
Loss & RA$\uparrow$ & IS$\uparrow$ & FID$\downarrow$ \\
\midrule
w/ $\mathcal{L}_{all}$ & $81.9\%$ & $58.172$ & $187.365$ \\
w/o $\mathcal{L}_{adv}$ & $0.0\%$  &$1.098$ &$287.104$\\
w/o $\mathcal{L}_{cls}$ & $0.1\%$ & $1.601$ & $318.286$\\
w/o $\mathcal{L}_{str}$ & $84.4\%$ & $57.578$ & $181.738$\\
w/o $\mathcal{L}_{sty}$ & $67.4\%$ & $55.108$ & $204.016$\\
w/o $\mathcal{L}_{rec}$ & $72.3\%$ & $54.316$ & $202.782$\\
\bottomrule
\end{tabular}
\end{table}

\begin{figure}[htbp]
    \centering
    \includegraphics[width=0.9\linewidth]{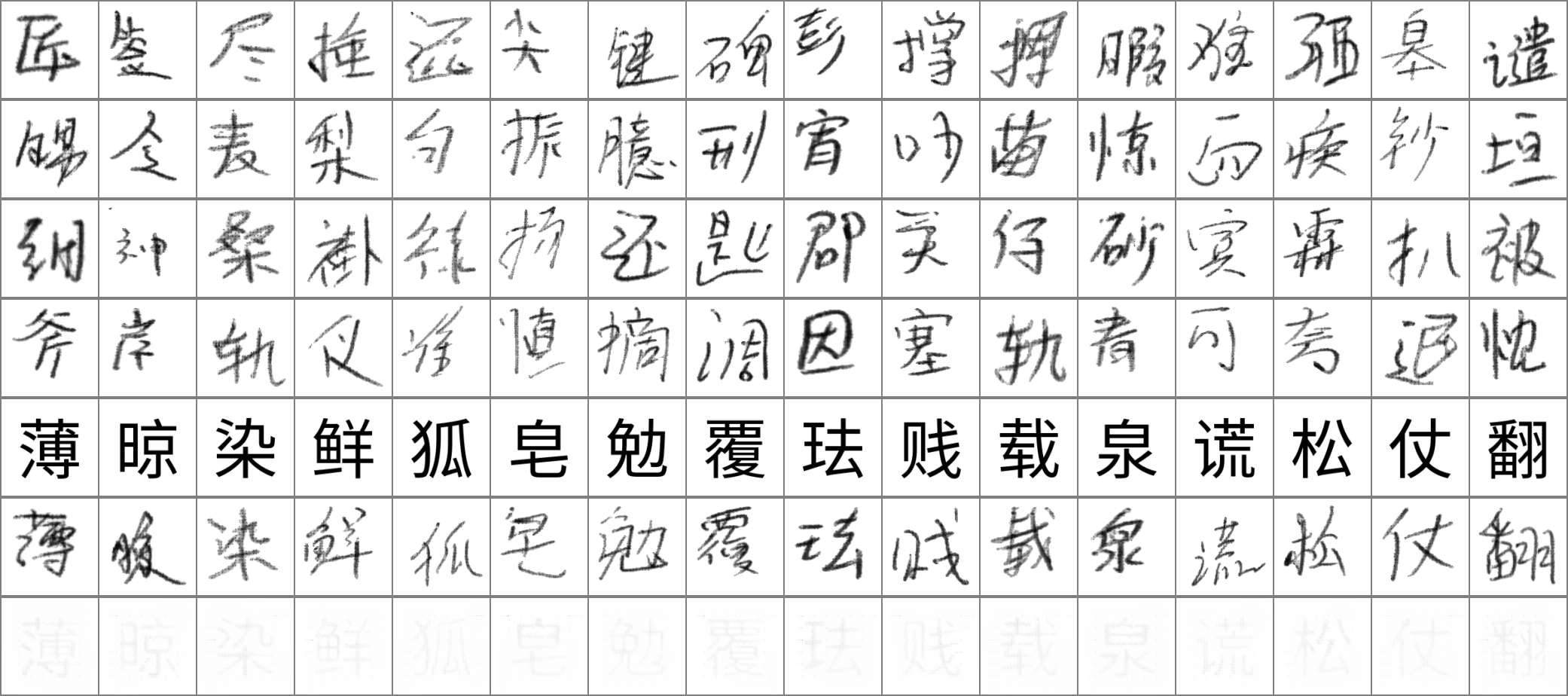}
    \caption{Examples of the generated results with $\mathcal{L}_{adv}$ removed.}
    \label{fig:ablation_adversarial}
\end{figure}

\begin{figure}[htbp]
    \centering
    \includegraphics[width=0.9\linewidth]{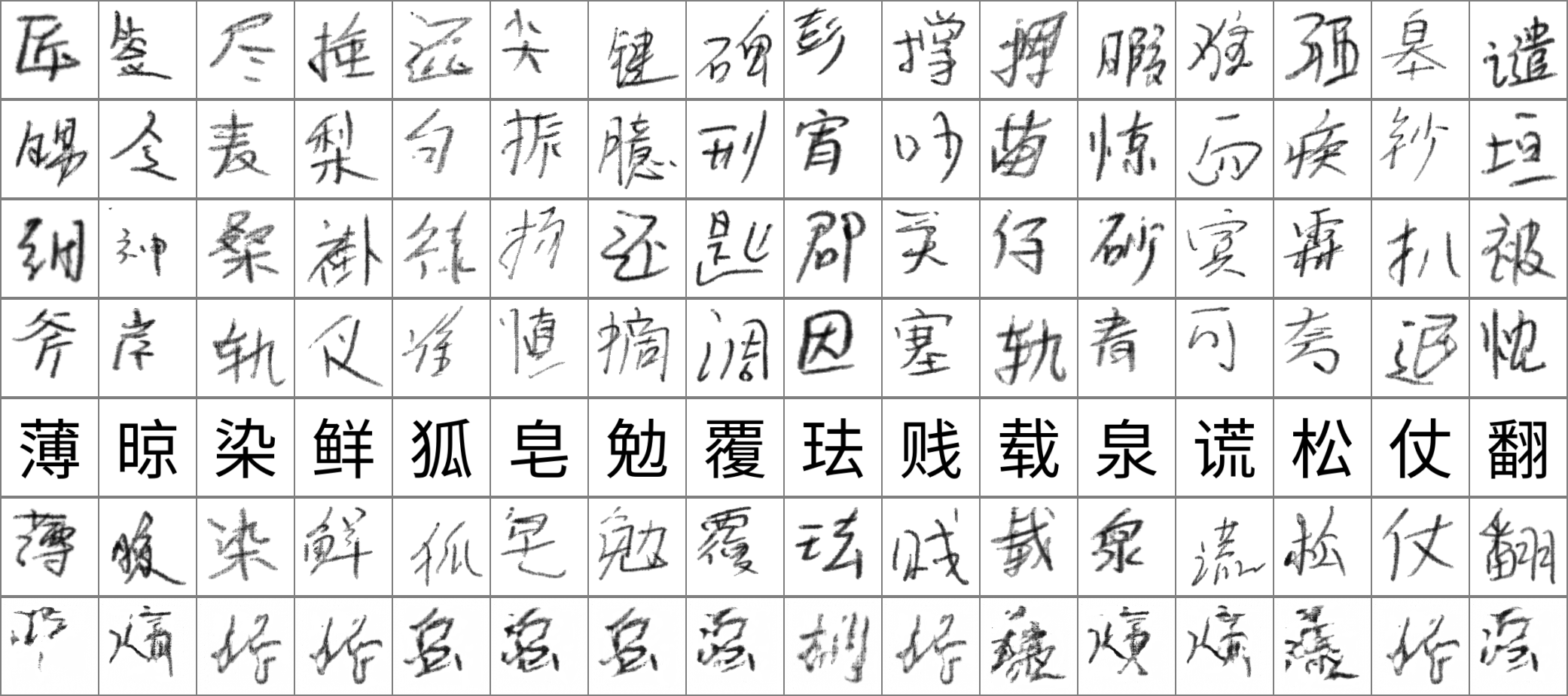}
    \caption{Examples of the generated results with $\mathcal{L}_{cls}$ removed.}
    \label{fig:ablation_classification}
\end{figure}

The models with one of the loss terms removed show performance degradation to varying degrees except for $\mathcal{L}_{str}$, which has been implicitly covered by the strong representation ability of the structure encoder. Removing $\mathcal{L}_{sty}$ and $\mathcal{L}_{rec}$ leads to a modest decrease in performance, while removing $\mathcal{L}_{adv}$ and $\mathcal{L}_{cls}$ results in a complete failure. We show some examples of the generated results with $\mathcal{L}_{adv}$ and $\mathcal{L}_{cls}$ removed in Figure \ref{fig:ablation_adversarial} and \ref{fig:ablation_classification} respectively. We can see that the generator cannot learn the handwritten style without adversarial learning, and fails to generate the correct character pattern with classification loss removed, which can even lead to serious mode collapse. Therefore, we claim that all the loss terms are necessary.
\section{Conclusion}
\label{sec:conclusion}
In this paper, we introduced MetaScript, a novel system for generating handwritten Chinese content using few-shot learning and Generative Adversarial Networks. Our approach effectively bridges the gap between the personalized nuances of handwriting and the efficiency of digital text generation. The key contributions of our work include the development of an innovative few-shot learning model, the integration of structural and stylistic elements in character generation, and the scalability and efficiency of the MetaScript system.

Our experiments demonstrate that MetaScript can successfully replicate a variety of handwriting styles with high fidelity using only a few style references. The system shows promising results in terms of recognition accuracy, inception score, and Frechet inception distance, indicating its effectiveness in generating authentic and diverse handwritten Chinese characters.

However, there are still challenges and limitations to be addressed. The quality of generated characters can vary depending on the number and quality of style references provided. Additionally, while our model performs well with common Chinese characters, its effectiveness with less common or more complex characters requires further exploration.

Future work will focus on enhancing the robustness and versatility of MetaScript: 1) We aim to enhance the robustness and versatility of the system, focusing on more sophisticated few-shot learning techniques. This enhancement is expected to significantly improve MetaScript's ability to learn effectively from limited data. 2) Another pivotal area of interest is the extension of our approach to non-Latin scripts, including Arabic and Devanagari. These scripts, with their rich handwriting traditions, present unique challenges and opportunities for our handwriting generation model. 3) Finally, we plan to integrate MetaScript into real-world applications. This integration involves embedding our system into digital education tools and personalized digital communication platforms, thereby infusing the warmth and personality of traditional handwriting into the digital realm.

{
    \small
    \bibliographystyle{ieeenat_fullname}
    \bibliography{main}

\begin{thebibliography}{46}
\providecommand{\natexlab}[1]{#1}
\providecommand{\url}[1]{\texttt{#1}}
\expandafter\ifx\csname urlstyle\endcsname\relax
  \providecommand{\doi}[1]{doi: #1}\else
  \providecommand{\doi}{doi: \begingroup \urlstyle{rm}\Url}\fi

\bibitem[Alonso et~al.(2019)Alonso, Moysset, and Messina]{ganhan1}
Eloi Alonso, Bastien Moysset, and Ronaldo Messina.
\newblock Adversarial generation of handwritten text images conditioned on sequences.
\newblock In \emph{2019 international conference on document analysis and recognition (ICDAR)}, pages 481--486. IEEE, 2019.

\bibitem[Bodapati et~al.(2020)Bodapati, Reddy, and Katta]{rehan3}
Suraj Bodapati, Sneha Reddy, and Sugamya Katta.
\newblock Realistic handwriting generation using recurrent neural networks and long short-term networks.
\newblock In \emph{Proceedings of the Third International Conference on Computational Intelligence and Informatics: ICCII 2018}, pages 651--661. Springer, 2020.

\bibitem[Chang et~al.(2018{\natexlab{a}})Chang, Zhang, Pan, and Meng]{chigan2}
Bo Chang, Qiong Zhang, Shenyi Pan, and Lili Meng.
\newblock Generating handwritten chinese characters using cyclegan.
\newblock In \emph{2018 IEEE winter conference on applications of computer vision (WACV)}, pages 199--207. IEEE, 2018{\natexlab{a}}.

\bibitem[Chang et~al.(2018{\natexlab{b}})Chang, Gu, Zhang, Wang, and Innovation]{chigan3}
Jie Chang, Yujun Gu, Ya Zhang, Yan-Feng Wang, and CM Innovation.
\newblock Chinese handwriting imitation with hierarchical generative adversarial network.
\newblock In \emph{BMVC}, page 290, 2018{\natexlab{b}}.

\bibitem[Chang(1973)]{chigenori}
Shi-Kuo Chang.
\newblock An interactive system for chinese character generation and retrieval.
\newblock \emph{IEEE Transactions on Systems, Man, and Cybernetics}, SMC-3\penalty0 (3):\penalty0 257--265, 1973.

\bibitem[Ding et~al.(2019)Ding, Liu, Yin, and Kong]{tgan}
Zihan Ding, Xiao-Yang Liu, Miao Yin, and Linghe Kong.
\newblock Tgan: Deep tensor generative adversarial nets for large image generation.
\newblock \emph{arXiv preprint arXiv:1901.09953}, 2019.

\bibitem[Durugkar et~al.(2016)Durugkar, Gemp, and Mahadevan]{gman}
Ishan Durugkar, Ian Gemp, and Sridhar Mahadevan.
\newblock Generative multi-adversarial networks.
\newblock \emph{arXiv preprint arXiv:1611.01673}, 2016.

\bibitem[Ehsani et~al.(2018)Ehsani, Mottaghi, and Farhadi]{sgan}
Kiana Ehsani, Roozbeh Mottaghi, and Ali Farhadi.
\newblock Segan: Segmenting and generating the invisible.
\newblock In \emph{Proceedings of the IEEE conference on computer vision and pattern recognition}, pages 6144--6153, 2018.

\bibitem[Fogel et~al.(2020)Fogel, Averbuch-Elor, Cohen, Mazor, and Litman]{ganhan2}
Sharon Fogel, Hadar Averbuch-Elor, Sarel Cohen, Shai Mazor, and Roee Litman.
\newblock Scrabblegan: Semi-supervised varying length handwritten text generation, 2020.

\bibitem[Gangadhar et~al.(2007)Gangadhar, Joseph, and Chakravarthy]{2007}
Garipelli Gangadhar, Denny Joseph, and V~Srinivasa Chakravarthy.
\newblock An oscillatory neuromotor model of handwriting generation.
\newblock \emph{International journal of document analysis and recognition (ijdar)}, 10:\penalty0 69--84, 2007.

\bibitem[Ghosh et~al.(2018)Ghosh, Kulharia, Namboodiri, Torr, and Dokania]{madgan}
Arnab Ghosh, Viveka Kulharia, Vinay~P Namboodiri, Philip~HS Torr, and Puneet~K Dokania.
\newblock Multi-agent diverse generative adversarial networks.
\newblock In \emph{Proceedings of the IEEE conference on computer vision and pattern recognition}, pages 8513--8521, 2018.

\bibitem[Graves(2013)]{rehan2}
Alex Graves.
\newblock Generating sequences with recurrent neural networks.
\newblock \emph{arXiv preprint arXiv:1308.0850}, 2013.

\bibitem[Gui et~al.(2023)Gui, Chen, Ding, and Huo]{giffchi2}
Dongnan Gui, Kai Chen, Haisong Ding, and Qiang Huo.
\newblock Zero-shot generation of training data with denoising diffusion probabilistic model for handwritten chinese character recognition.
\newblock \emph{arXiv preprint arXiv:2305.15660}, 2023.

\bibitem[Gui et~al.(2020)Gui, Sun, Wen, Tao, and Ye]{ganreview}
Jie Gui, Zhenan Sun, Yonggang Wen, Dacheng Tao, and Jieping Ye.
\newblock A review on generative adversarial networks: Algorithms, theory, and applications, 2020.

\bibitem[He et~al.(2022)He, Chen, Wang, Liu, Du, Tao, and Qiao]{giffchi1}
Haibin He, Xinyuan Chen, Chaoyue Wang, Juhua Liu, Bo Du, Dacheng Tao, and Yu Qiao.
\newblock Diff-font: Diffusion model for robust one-shot font generation.
\newblock \emph{arXiv preprint arXiv:2212.05895}, 2022.

\bibitem[He et~al.(2016)He, Zhang, Ren, and Sun]{resnet}
Kaiming He, Xiangyu Zhang, Shaoqing Ren, and Jian Sun.
\newblock Deep residual learning for image recognition.
\newblock In \emph{Proceedings of the IEEE conference on computer vision and pattern recognition}, pages 770--778, 2016.

\bibitem[Heusel et~al.(2017)Heusel, Ramsauer, Unterthiner, Nessler, and Hochreiter]{fid}
Martin Heusel, Hubert Ramsauer, Thomas Unterthiner, Bernhard Nessler, and Sepp Hochreiter.
\newblock Gans trained by a two time-scale update rule converge to a local nash equilibrium.
\newblock \emph{Advances in neural information processing systems}, 30, 2017.

\bibitem[Ji et~al.(2022)Ji, Wu, Hu, He, Chen, and He]{personality}
Yu Ji, Wen Wu, Yi Hu, Xiaofeng He, Changzhi Chen, and Liang He.
\newblock Automatic personality prediction based on users’ chinese handwriting change.
\newblock In \emph{CCF Conference on Computer Supported Cooperative Work and Social Computing}, pages 435--449. Springer, 2022.

\bibitem[Jiang et~al.(2018)Jiang, Yang, Huang, and Zhang]{wnet}
Haochuan Jiang, Guanyu Yang, Kaizhu Huang, and Rui Zhang.
\newblock W-net: One-shot arbitrary-style chinese character generation with deep neural networks.
\newblock In \emph{Neural Information Processing}, pages 483--493, Cham, 2018. Springer International Publishing.

\bibitem[Jolicoeur-Martineau(2018)]{rgans}
Alexia Jolicoeur-Martineau.
\newblock The relativistic discriminator: a key element missing from standard gan.
\newblock \emph{arXiv preprint arXiv:1807.00734}, 2018.

\bibitem[Kanda et~al.(2020)Kanda, Iwana, and Uchida]{rlhan}
Keisuke Kanda, Brian~Kenji Iwana, and Seiichi Uchida.
\newblock What is the reward for handwriting? — a handwriting generation model based on imitation learning.
\newblock In \emph{2020 17th International Conference on Frontiers in Handwriting Recognition (ICFHR)}, pages 109--114, 2020.

\bibitem[Karras et~al.(2019)Karras, Laine, and Aila]{stylegan}
Tero Karras, Samuli Laine, and Timo Aila.
\newblock A style-based generator architecture for generative adversarial networks.
\newblock In \emph{Proceedings of the IEEE/CVF conference on computer vision and pattern recognition}, pages 4401--4410, 2019.

\bibitem[Karras et~al.(2020)Karras, Laine, Aittala, Hellsten, Lehtinen, and Aila]{stylegan2}
Tero Karras, Samuli Laine, Miika Aittala, Janne Hellsten, Jaakko Lehtinen, and Timo Aila.
\newblock Analyzing and improving the image quality of stylegan.
\newblock In \emph{Proceedings of the IEEE/CVF conference on computer vision and pattern recognition}, pages 8110--8119, 2020.

\bibitem[Kumar et~al.(2018)Kumar, Kandala, and Reddy]{rehan1}
K.~Manoj Kumar, Harish Kandala, and N.~Sudhakar Reddy.
\newblock Synthesizing and imitating handwriting using deep recurrent neural networks and mixture density networks.
\newblock In \emph{2018 9th International Conference on Computing, Communication and Networking Technologies (ICCCNT)}, pages 1--6, 2018.

\bibitem[Ledig et~al.(2017)Ledig, Theis, Husz{\'a}r, Caballero, Cunningham, Acosta, Aitken, Tejani, Totz, Wang, et~al.]{ganphoto}
Christian Ledig, Lucas Theis, Ferenc Husz{\'a}r, Jose Caballero, Andrew Cunningham, Alejandro Acosta, Andrew Aitken, Alykhan Tejani, Johannes Totz, Zehan Wang, et~al.
\newblock Photo-realistic single image super-resolution using a generative adversarial network.
\newblock In \emph{Proceedings of the IEEE conference on computer vision and pattern recognition}, pages 4681--4690, 2017.

\bibitem[Li et~al.(2019{\natexlab{a}})Li, Bao, Yang, Chen, and Wen]{faceshifter}
Lingzhi Li, Jianmin Bao, Hao Yang, Dong Chen, and Fang Wen.
\newblock Faceshifter: Towards high fidelity and occlusion aware face swapping.
\newblock \emph{arXiv preprint arXiv:1912.13457}, 2019{\natexlab{a}}.

\bibitem[Li et~al.(2019{\natexlab{b}})Li, Wang, Yang, Huang, and Du]{chigan1}
Meng Li, Jian Wang, Yi Yang, Weixing Huang, and Wenjuan Du.
\newblock Improving gan-based calligraphy character generation using graph matching.
\newblock In \emph{2019 IEEE 19th International Conference on Software Quality, Reliability and Security Companion (QRS-C)}, pages 291--295. IEEE, 2019{\natexlab{b}}.

\bibitem[Liao et~al.(2023)Liao, Xia, and Wang]{giffchi3}
Qisheng Liao, Gus Xia, and Zhinuo Wang.
\newblock Calliffusion: Chinese calligraphy generation and style transfer with diffusion modeling.
\newblock \emph{arXiv preprint arXiv:2305.19124}, 2023.

\bibitem[Liu et~al.(2011)Liu, Yin, Wang, and Wang]{casia}
Cheng-Lin Liu, Fei Yin, Da-Han Wang, and Qiu-Feng Wang.
\newblock Casia online and offline chinese handwriting databases.
\newblock In \emph{2011 international conference on document analysis and recognition}, pages 37--41. IEEE, 2011.

\bibitem[Liu and Tuzel(2016)]{cogan}
Ming-Yu Liu and Oncel Tuzel.
\newblock Coupled generative adversarial networks.
\newblock \emph{Advances in neural information processing systems}, 29, 2016.

\bibitem[Liu et~al.(2021)Liu, Meng, Xiang, and Pan]{ganhan3}
Xiyan Liu, Gaofeng Meng, Shiming Xiang, and Chunhong Pan.
\newblock Handwritten text generation via disentangled representations.
\newblock \emph{IEEE Signal Processing Letters}, 28:\penalty0 1838--1842, 2021.

\bibitem[Lu et~al.(2018)Lu, Tai, and Tang]{cyclegan}
Yongyi Lu, Yu-Wing Tai, and Chi-Keung Tang.
\newblock Attribute-guided face generation using conditional cyclegan.
\newblock In \emph{Proceedings of the European conference on computer vision (ECCV)}, pages 282--297, 2018.

\bibitem[Madaan et~al.(2022)Madaan, Kumar, Kumar, Saha, and Gupta]{hanreview}
Mehul Madaan, Aniket Kumar, Shubham Kumar, Aniket Saha, and Kirti Gupta.
\newblock Handwriting generation and synthesis: A review.
\newblock In \emph{2022 Second International Conference on Power, Control and Computing Technologies (ICPC2T)}, pages 1--6, 2022.

\bibitem[Mogren(2016)]{mogren}
Olof Mogren.
\newblock C-rnn-gan: Continuous recurrent neural networks with adversarial training.
\newblock \emph{arXiv preprint arXiv:1611.09904}, 2016.

\bibitem[Ratliff et~al.(2013)Ratliff, Burden, and Sastry]{nash}
Lillian~J Ratliff, Samuel~A Burden, and S~Shankar Sastry.
\newblock Characterization and computation of local nash equilibria in continuous games.
\newblock In \emph{2013 51st Annual Allerton Conference on Communication, Control, and Computing (Allerton)}, pages 917--924. IEEE, 2013.

\bibitem[Ronneberger et~al.(2015)Ronneberger, Fischer, and Brox]{unet}
Olaf Ronneberger, Philipp Fischer, and Thomas Brox.
\newblock U-net: Convolutional networks for biomedical image segmentation.
\newblock In \emph{Medical Image Computing and Computer-Assisted Intervention--MICCAI 2015: 18th International Conference, Munich, Germany, October 5-9, 2015, Proceedings, Part III 18}, pages 234--241. Springer, 2015.

\bibitem[Salimans et~al.(2016)Salimans, Goodfellow, Zaremba, Cheung, Radford, and Chen]{is}
Tim Salimans, Ian Goodfellow, Wojciech Zaremba, Vicki Cheung, Alec Radford, and Xi Chen.
\newblock Improved techniques for training gans.
\newblock \emph{Advances in neural information processing systems}, 29, 2016.

\bibitem[Tan and Le(2021)]{efficientnetv2}
Mingxing Tan and Quoc Le.
\newblock Efficientnetv2: Smaller models and faster training.
\newblock In \emph{International conference on machine learning}, pages 10096--10106. PMLR, 2021.

\bibitem[Vondrick et~al.(2016)Vondrick, Pirsiavash, and Torralba]{vgan}
Carl Vondrick, Hamed Pirsiavash, and Antonio Torralba.
\newblock Generating videos with scene dynamics.
\newblock \emph{Advances in neural information processing systems}, 29, 2016.

\bibitem[Wang et~al.(2018{\natexlab{a}})Wang, Liu, Zhu, Tao, Kautz, and Catanzaro]{pix2pixhd}
Ting-Chun Wang, Ming-Yu Liu, Jun-Yan Zhu, Andrew Tao, Jan Kautz, and Bryan Catanzaro.
\newblock High-resolution image synthesis and semantic manipulation with conditional gans.
\newblock In \emph{Proceedings of the IEEE conference on computer vision and pattern recognition}, pages 8798--8807, 2018{\natexlab{a}}.

\bibitem[Wang et~al.(2018{\natexlab{b}})Wang, Yu, Wu, Gu, Liu, Dong, Qiao, and Change~Loy]{esrgan}
Xintao Wang, Ke Yu, Shixiang Wu, Jinjin Gu, Yihao Liu, Chao Dong, Yu Qiao, and Chen Change~Loy.
\newblock Esrgan: Enhanced super-resolution generative adversarial networks.
\newblock In \emph{Proceedings of the European conference on computer vision (ECCV) workshops}, pages 0--0, 2018{\natexlab{b}}.

\bibitem[Xu et~al.(2009)Xu, Jin, Jiang, and Lau]{chier1}
Songhua Xu, Tao Jin, Hao Jiang, and Francis~CM Lau.
\newblock Automatic generation of personal chinese handwriting by capturing the characteristics of personal handwriting.
\newblock In \emph{Twenty-First IAAI Conference}, 2009.

\bibitem[Yu et~al.(2017)Yu, Zhang, Wang, and Yu]{seqgan}
Lantao Yu, Weinan Zhang, Jun Wang, and Yong Yu.
\newblock Seqgan: Sequence generative adversarial nets with policy gradient.
\newblock In \emph{Proceedings of the AAAI conference on artificial intelligence}, 2017.

\bibitem[Yuan et~al.(2018)Yuan, Liu, Zhang, Zhang, Dong, and Lin]{cicgan}
Yuan Yuan, Siyuan Liu, Jiawei Zhang, Yongbing Zhang, Chao Dong, and Liang Lin.
\newblock Unsupervised image super-resolution using cycle-in-cycle generative adversarial networks.
\newblock In \emph{Proceedings of the IEEE conference on computer vision and pattern recognition workshops}, pages 701--710, 2018.

\bibitem[Zhang et~al.(2018)Zhang, Qiu, Yao, Liu, and Mei]{fcan}
Yiheng Zhang, Zhaofan Qiu, Ting Yao, Dong Liu, and Tao Mei.
\newblock Fully convolutional adaptation networks for semantic segmentation.
\newblock In \emph{Proceedings of the IEEE conference on computer vision and pattern recognition}, pages 6810--6818, 2018.

\bibitem[Zhou et~al.(2011)Zhou, Wang, and Chen]{chier2}
Baoyao Zhou, Weihong Wang, and Zhanghui Chen.
\newblock Easy generation of personal chinese handwritten fonts.
\newblock In \emph{2011 IEEE international conference on multimedia and expo}, pages 1--6. IEEE, 2011.

\end{thebibliography}
}

\end{document}